# Fast-AT: Fast Automatic Thumbnail Generation using Deep Neural Networks

Seyed A. Esmaeili    Bharat Singh    Larry S. Davis

University of Maryland, College Park

{sesmaeil@umd.edu, bharat@cs.umd.edu, lsd@umiacs.umd.edu}

## Abstract

*Fast-AT is an automatic thumbnail generation system based on deep neural networks. It is a fully-convolutional deep neural network, which learns specific filters for thumbnails of different sizes and aspect ratios. During inference, the appropriate filter is selected depending on the dimensions of the target thumbnail. Unlike most previous work, Fast-AT does not utilize saliency but addresses the problem directly. In addition, it eliminates the need to conduct region search on the saliency map. The model generalizes to thumbnails of different sizes including those with extreme aspect ratios and can generate thumbnails in real time. A data set of more than 70,000 thumbnail annotations was collected to train Fast-AT. We show competitive results in comparison to existing techniques.*

## 1. Introduction

Thumbnails are used to facilitate browsing of a collection of images, make economic use of display space, and reduce the transmission time. A thumbnail image is a smaller version of an original images that is meant to effectively portray the original image (Figure 1). Social media websites such as Facebook, Twitter, Pinterest, etc have content from multiple user accounts which needs to be displayed on a fixed resolution display. A normal web page on Facebook contains hundreds of images, which are essentially thumbnails of larger images. Therefore, it is important to ensure that each thumbnail displays the most useful information present in the original image. Since images displayed on a web page vary significantly in *size* and *aspect ratio*, any thumbnail generation algorithm must be able to generate thumbnails over a range of scales and aspect ratios.

The standard operations used for creating thumbnails are cropping and scaling. Since thumbnails are ubiquitous and the manual generation of thumbnails is time consuming, significant research has been devoted for automatic thumbnail generation. Most methods [20, 3, 2] utilize a saliency map to identify regions in the image that could serve as good crops to create thumbnails. This leads to a two step solution where saliency is first computed and then an optimization problem is solved to find the optimum crop. Whereas a recent method addresses [9] the problem directly, it involves hand crafted features and uses SVMs to score candidate crops. Moreover, the implementation requires 60 seconds to produce a single thumbnail.

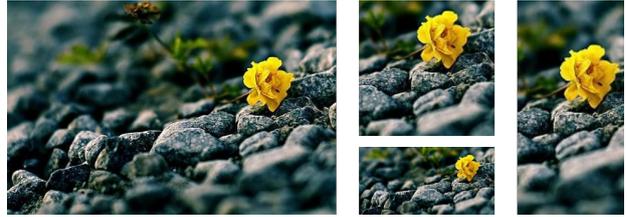

Figure 1. Illustration of the thumbnail problem. The original image is shown on the left with thumbnails of different aspect ratios on the right.

We propose Fast-AT, a deep learning based approach for thumbnail generation that addresses the problem directly, in an end-to-end learning framework. Our work involves the following contributions:

- Fast-AT is based on an object detection framework, which takes dimensions of the target thumbnail into account for generating crops. Since it produces thumbnails using a feed-forward network, it can process 9 images per second on a GPU.

- By vector quantizing aspect ratios, Fast-AT learns different filters for different aspect ratios during training. During inference, the appropriate filter is selected depending on dimensions of the target thumbnail.

- 70,048 thumbnail annotations were created on 28,064 images through Amazon Mechanical Turk. The annotated thumbnail data set will be released with this paper.

## 2. Related Work

Since thumbnail creation involves reducing the image size, retargetting methods [17] such as seam carving and non-homogeneous warping can be used. However these methods produce artifacts which are often pronounced since most thumbnails are significantly smaller than the original images. Therefore, most thumbnail generation methods use a combination of cropping and scaling. Typically, automatic thumbnail generators utilize a saliency map as an indicator of important regions in the image to be cropped [20, 3, 21, 2]. Region search is then performed to find the smallest region of the image that has a total saliency above a certain threshold. A brute force approach to region search is computationally expensive, therefore approximations have been investigated such as greedy search [20], restriction of the search space to a set of fixed size rectangles [19], and binarization of the saliency map [3]. Recently, an algorithm that conducts region search in linear time has been reported [2].

However, saliency can ignore the semantics of the scene and does not take the target thumbnail size into account. Many methods address this shortcoming through a heuristic approach such as selecting a crop that contains all detected faces [20] or using an algorithm that depends on both the saliency and image class. Sun et al. [21] improve the saliency map by taking the thumbnail scale into account and preserving object completeness. A scale and object aware saliency map is computed by using a contrast sensitivity function [14] and an objectiveness measure [1]; then greedy search is conducted to find the optimum region, similar to [20]. However, the method does not impose aspect ratio restrictions on the selected region and the final thumbnail images can contain objects that look significantly deformed. In addition, whereas [2] introduced an algorithm which produces regions with restricted aspect ratios, it mentions that the problem could be infeasible for a given overall saliency threshold value. Some other approaches attempt to crop the most aesthetic part of the image [23, 15].

Huang et al. were the first to directly address the problem [9]. A data set of images and their manually generated thumbnails was collected. However, only one thumbnail size of 160×120 was considered. The solution also involved scoring a large set of candidate crops and then selecting the crop with the largest score. The implementation - albeit an unoptimized CPU code - required 60 seconds to generate a thumbnail for a single image. In addition, the solution [9] is based on hand crafted features and SVM, which generally have inferior performance compared to recent deep learning based methods.

Deep convolutional neural networks have achieved impressive results on high level visual tasks such as image classification [11, 18, 8], object detection [7, 16, 4], and semantic segmentation[12]. These architectures have not only led to significantly better results but also systems that can be deployed in real time [16]. We present a solution, based on a fully convolutional deep neural network that is learned end-to-end. We take into account varying thumbnail sizes from 32×32 to 200×200 pixels. At test time the network can produce thumbnails at 105ms per image and shows significant improvements over the existing baselines.

## 3. Data Collection

We started the annotation of images from the photo quality data set of [13] using Amazon Mechanical Turk (AMT). The set includes both high and low quality images and spans a number of categories such as humans, animals, and landscape. Target thumbnail sizes were divided into three groups - thumbnails between 32 to 64, 64 to 128, and 100 to 200, in both height and width. This leads to an aspect ratio range from 0.5 to 2. Each image was annotated three times, with different target thumbnail sizes from each group.

The annotation was done through an interface that draws a bounding box on the original image with an aspect ratio equal to that of the thumbnail; users can only scale the box up or down and change its location. This bounding box represents the selected crop. It is scaled down to the thumbnail size and shown to the user at the same time. Restricting the bounding box (crop) to have an aspect ratio equal to the thumbnail's aspect ratio leads to more flexible annotation and avoids any possible deformation affects.

To make the interface more practical, the images were scaled down such that the height does not exceed 650 and the width does not exceed 800. The Mechanical Turk workers were shown examples of good and bad thumbnails. The examples were intended to illustrate that good thumbnails capture a significant amount of content while at the same time are easy to recognize. After the data set was collected, the thumbnail images were manually swept through and bad annotations were excluded; this led to a total of 70,048 annotations over 28,064 images with each image having at most 3 annotations.

## 4. Does target thumbnail size matter?

An automatic thumbnail generating system receives two inputs: the image and the target thumbnail. Therefore, we study the dependence between the target thumbnail and the generated crop. It is clear that the generated crop should have an aspect ratio equal to the target thumbnail's aspect ratio. Selecting a crop of a different aspect ratio could cause pronounced deformations when scaling the crop down to the thumbnail size as shown in Figure 2. It is worth noting that although deformations can be caused when selecting crops with aspect ratios different from that of the thumbnail, it has been ignored in some work [20, 21].

Intuitively, it is expected that smaller input thumbnail sizes would typically require smaller crops. Larger crops would be less recognizable when they are scaled down. To investigate this, we compare the average area of the anno-tated crop vs the thumbnail size in the annotated dataset. However, we did not observe any correlation between the two as shown in Figure 3. Thus, we concluded that to produce the optimal crop, the thumbnail size does not need to be taken into account, but the aspect ratio matters for this dataset. We still consider a model that takes both the aspect ratio and the size of the thumbnail into account in our experiments.

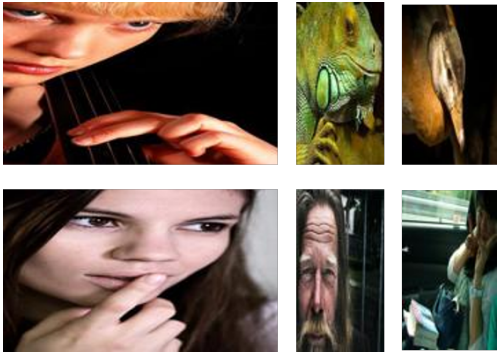

Figure 2. The above thumbnails were generated using the code from [21] which is agnostic to the thumbnail aspect ratio. The objects look clearly deformed in the thumbnails, where variation in aspect ratio is significant.

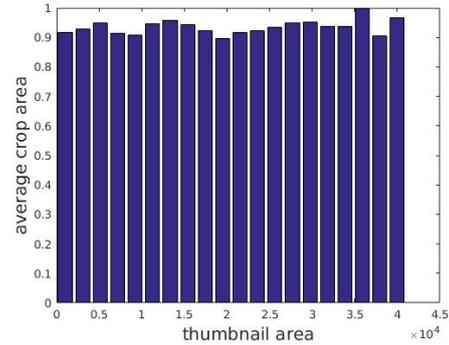

Figure 3. The plot above shows the average crop area for a given thumbnail size. The crop area is normalized by the maximum value. The average crop area is not generally smaller for smaller thumbnail sizes and it does not vary significantly.

## 5. Approach

Thumbnails are created by selecting a region in the image to be cropped (a bounding box) followed by scaling the bounding box down to the thumbnail size. We present a solution to this problem employing a deep convolutional neural network that learns the best bounding boxes to produce thumbnails. Since we formulate the problem as a bounding box prediction problem, it is closely connected to object detection. However, unlike object detection, the final predictions will not consist of bounding boxes with a discrete probability distribution across different classes, but involve two classes: one that is representative of the image and another which is not.

Early deep learning methods for object detection utilized proposal methods such as [22] that were time-consuming. Computation time was significantly reduced using region proposal networks (RPNs) [16] that learn to generate proposals. R-FCN [4], which was recently proposed, reduces the computational expense of forward propagating the pooled proposal features through two fully connected layers by introducing a new convolutional layer consisting of class-specific position-sensitive filters. Specifically, if there are $C$ classes to be detected, then this new convolutional layer will generate $k^2(C+1)$ feature maps. The $k^2$ position-sensitive score maps correspond to $k \times k$ evenly partitioned cells. Those $k^2$ feature maps are associated with different relative positions in the spatial gird, such as (top-left,...,bottom-left) for every class. $k = 3$, corresponds to a $3 \times 3$ spatial grid and 9 position-sensitive filters per class. Every class (including the background), will have $k^2$ feature maps associated with it. Instead of forward propagating through two fully connected layers, position-sensitive pooling followed by score averaging is performed. This generates a $(C+1)$-d vector on which a softmax function is applied to obtain responses across categories.

An architecture for thumbnail generation should be fully convolutional because, including a fully connected layer requires a fixed input size. If there were a mismatch between the aspect ratio of an image and the fixed input size, the image would have to be cropped in addition to being scaled. Because the thumbnail crops (bounding boxes) could touch the boundaries of the image or even extend to the whole image, the pre-processing step of cropping a region of the image could lead to sub-optimal predictions, since part of the image has been removed. Therefore, an architecture similar to [18] that was used for the ImageNet localization challenge [5], which simply replaces the class scores by 4-D bounding predictions, cannot be employed because of the fully connected layer at the end.

Another observation is that unlike object detection, the thumbnail generation network receives two inputs: the image and the thumbnail aspect ratio. Both RPN and R-FCN introduce task-specific filters. In the case of RPN, filter banks that specialize in predicting proposals of a specific scale are achieved by modifying the training policy. In the case of R-FCN, position-sensitive filters specialize through the position-sensitive pooling mechanism. In a similar manner, we modify R-FCN for thumbnail creation by introducing a set of aspect ratio-specific filter banks. A set of $A$ points are introduced in the aspect ratio range of $[0.5, 2]$, which represent aspect ratios that grow by a constant fac-tor (a geometric sequence), i.e. it is of the form $S = \{\frac{1}{2}c, \frac{1}{2}c^2, \ldots, \frac{1}{2}c^A\}$. Note that $\frac{1}{2}c^0 = \frac{1}{2}$ and $\frac{1}{2}c^{(A+1)} = 2$, leading to $c = \sqrt[A+1]{4}$. The filter banks in the last convolutional layer in R-FCN are modified into $A$ pairs, with each pair having a total of $k^2$ filters. Each pair is associated with a single element in the set $S$. Similar to R-FCN, position-sensitive pooling followed by averaging is performed over that pair and those two values are used to yield a softmax prediction of representativeness.

At training time, when an image-thumbnail size pair is received, the image is forward propagated through convolutional layers up to the last convolutional layer. The thumbnail aspect ratio is calculated and the element with the closest value is selected from $S$ - the pair associated with that element factors in the training while others are ignored. For this pair, the proposals are received, and similar to object detection, positive and negative labels are assigned to the proposals based on their intersection over union (IoU) with the ground truth. Specifically, a positive label is assigned if the IoU $\geq 0.5$ and negative otherwise. Similarly, $A$ aspect ratio-specific regressors are trained, one for each element in $S$; these are similar to class-specific regressors. For a given proposal, we employ the following loss:

$$L(s_i, t_i) = \sum_{i=1}^{A} l_i L_{cls}(s_i, s^*) + \lambda[s^* = 1 \wedge l_i] L_{reg}(t_i, t^*)$$

, where $l_i$ is either ignore=0 or factor-in=1, namely

$$l_i = \begin{cases} 1 & \text{if } i = \underset{i}{\operatorname{argmin}} |\frac{1}{2}c^i - \text{thumbnail aspect ratio}| \\ 0 & \text{otherwise} \end{cases}$$

$s_i$ is the representativeness score predicted by the $i^{th}$ pair, $s^*$ is the ground truth label, and $L_{cls}$ is cross entropy loss. $\lambda$ is a weight for the regression loss which we set to 1. The regression loss is 0 for all but the nearest aspect ratio. For the filter corresponding to the nearest aspect ratio, $L_{reg}$ is the smooth $L_1$ loss as defined in [6], $t_i$ is the bounding box predictions made by the $i^{th}$ regressor and $t^*$ is the ground truth bounding box. Both predictions are parametrized as in [6]. Figure 4 illustrates the architecture.

Since each regressor is responsible for a range of input thumbnail sizes, the predictions made by any regressor at test time could have an aspect ratio that differs from the target thumbnail aspect ratio. Therefore the output bounding box has to be rectified to have an aspect ratio equal to the thumbnail's aspect ratio, to eliminate any possible deformations when scaling down. We employ a simple method where a new bounding box with an aspect ratio equal to that of the target thumbnail is placed at the center of the predicted box and is expanded until it touches the boundaries. Since the predicted box already has an aspect ratio close to that of the thumbnail, the difference between the rectified

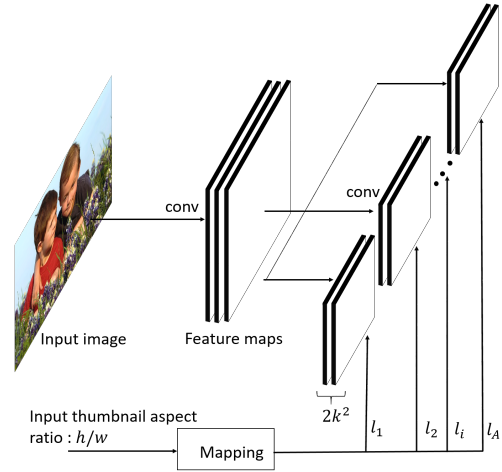

Figure 4. Illustration of the Fast-AT architecture and training policy. The appropriate filter is decided based on the thumbnail aspect ratio.

box and the predicted box is not significant, as shown in Figure 5(b).

Our implementation of Fast-AT is based on Resnet-101 [8], a learning rate of 0.001, momentum of 0.9, weight decay of 0.0005 with approximate joint training is used [16].

### 5.1. Does R-FCN alone work?

Among the baselines we consider is R-FCN- without any modifications. In effect, it is performing object detection between two classes. We find that R-FCN alone generates bounding boxes that have good representations of the original image. But since the architecture is agnostic to the input thumbnail dimensions, the generated thumbnails are of low quality as shown in Figure 5(a). If we apply the same rectification to the generated boxes, to cancel the deformation affects, important parts of the images are not preserved, in contrast to our model's results which are shown in Figure. 5(b). This is because of the significant mismatch between the target thumbnail aspect ratio and the predicted box aspect ratio. Eliminating the rectification step would lead to deformed results, similar to what is shown in Figure 2.

## 6. Experiments

In comparing models we use the following metrics:

- offset: the distance between the center of the ground truth bounding box and the predicted bounding box.

- rescaling factor (rescaling): defined as $max(s_g/s_p, s_p/s_g)$ where $s_g$ and $s_p$ are the rescaling factors for the ground truth and predicted box, respectively. [9].

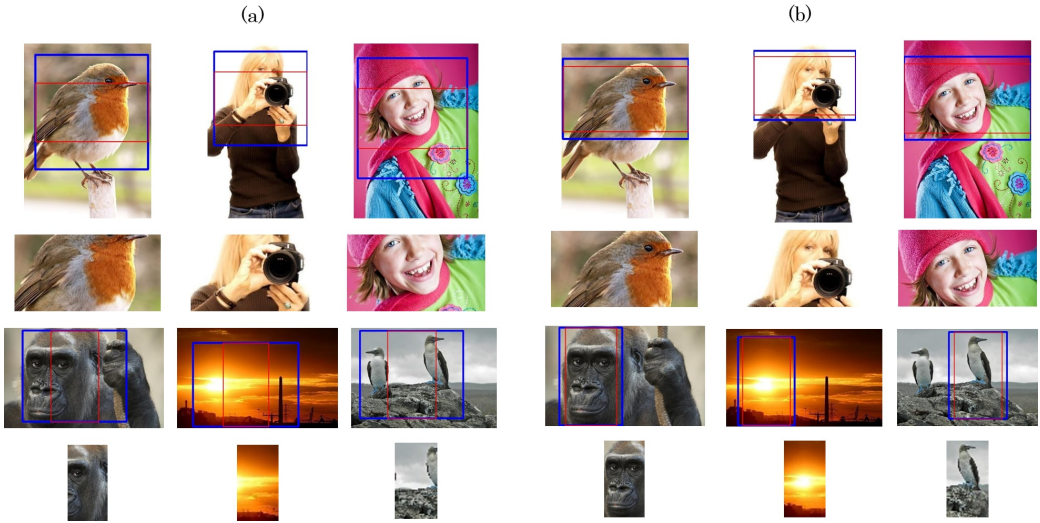

Figure 5. R-FCN and Fast-AT predictions on test set images. (a):The original image is displayed with the R-FCN prediction in blue, the rectified box is in red and the resulting thumbnail is below. Note how the resulting thumbnail is missing important parts of the original image.(b):The original image is displayed with the Fast-AT prediction in blue, the rectified box is in red and the resulting thumbnail is below. The rectification does not introduce a significant change in the box predicted by Fast-AT.

| **Model** | offset | rescaling | IoU | mismatch |
|---|---|---|---|---|
| R-FCN | 56.2 | 1.192 | 0.64 | 0.102 |
| Fast-AT (AR) | 55.0 | 1.149 | 0.68 | 0.010 |
| Fast-AT (AR+TS) | 55.4 | 1.154 | 0.68 | 0.012 |
| Fast-AT (AR, scale up of 350) | 53.1 | 1.156 | 0.69 | 0.024 |

Table 1. Metrics computed using different models. R-FCN, Fast-AT with aspect ratio mapping (AR), Fast-AT with aspect ratio and thumbnail size mapping (AR+TS), and Fast-AT with aspect ratio mapping scaled up to 350.

- IoU: intersection over union between the predicted box and the ground truth.

- aspect ratio mismatch (mismatch): the square of the difference between the aspect ratio of the predicted box and the aspect ratio of the thumbnail.

The total dataset consisting of 70,048 annotations over 28,064 images was split into 24,154 images for training with 63043 annotations (90% of the total annotations) and 3,910 images for testing with 7,005 annotations (10% of the total annotations). The training and test sets do not share any images. Comparative results between different models is shown in Table 1. The first model we use is R-FCN without modification. This architecture is agnostic to thumbnail dimension, the number of classes are reduced to two and the architecture is modified accordingly. We see that R-FCN alone has good performance in terms of all metrics except for aspect ratio mismatch. High values in these metrics show that rectifying the bounding box will cause significant change in the predicted box. We next consider our proposed model, where we map based on aspect ratio, with 5 divisions ($A = 5$). We see significant improvements in the metrics - the mean IoU has increased by 4% and the offset and rescaling factor have been reduced. The aspect ratio mismatch has also been significantly reduced. We extend the divisions to thumbnail sizes as well. In this case we divide the input thumbnail space into three branches: small thumbnails (32-64), medium (64-100), and large (100-200). Each branch is further divided based on aspect ratio as in the first model. This leads to a total of $5 \times 3 = 15$ regressors and 15 pairs of $2k^2$ filters. This did not lead to an improvement over the model with only aspect ratio divisions.

Unlike object detection bounding boxes, the predicted bounding boxes for thumbnails can enclose multiple objects and may extend to the whole image. So while, a network with a small receptive field may predict accurate bounding boxes for object detection, its predictions for thumbnail crops may be inaccurate. For object detection, Faster RCNN [16] effectively reduces the receptive field of the network by scaling up the image so that the smallest dimension is 600. This step is implemented in R-FCN as well. We reduce the image dimension (minimum of height/width) from 600 to 350 in Fast-AT to investigate the affect of the receptive field. We observe a slight improvement in the offset and IoU as shown in Table 1. The improvement is not large because we use Resnet-101 [8], which already has a large receptive field. Fast-AT's architecture is generic and hence other models such as VGGNet [18] or ZFNet [24] can also be used. Such shallower models are likely to benefit significantly in thumbnail generation if their receptive fields are extended.

We further compare Fast-AT with aspect ratio divisions and Fast-AT with aspect ratio and thumbnail divisions over thumbnails with a small size; below $64 \times 64$. We do not see any significant improvement (Table 2). This further confirms our initial conclusion that it is the thumbnail aspect ratio, not the thumbnail size, that matters.

| Model | offset | rescaling | IoU | mismatch |
|---|---|---|---|---|
| Fast-AT (AR) | 55.9 | 1.149 | 0.67 | 0.011 |
| Fast-AT (AR+TS) | 55.0 | 1.153 | 0.68 | 0.012 |

Table 2. Performance Comparison between Fast-AT with aspect ratio mapping (AR) and Fast-AT with aspect ratio and thumbnail size mapping (AR+TS) at small thumbnail sizes below $64 \times 64$.

We also measure the metrics at extreme aspect ratios, i.e. aspect ratios below 0.7 and above 1.8; the results are shown in Table 3. We observe a significant drop in R-FCN's performance- IoU falls by about 6% and the mismatch almost doubles. At the same time Fast-AT still performs well. This shows that Fast-AT can handle thumbnails of widely varying aspect ratios.

| Model | offset | rescaling | IoU | mismatch |
|---|---|---|---|---|
| R-FCN | 57.5 | 1.348 | 0.58 | 0.200 |
| Fast-AT (AR) | 49.8 | 1.18 | 0.68 | 0.013 |
| Fast-AT (AR+TS) | 50.7 | 1.183 | 0.68 | 0.014 |

Table 3. Performance Comparison between R-FCN, Fast-AT with aspect ratio mapping (AR) and Fast-AT with aspect ratio and thumbnail mapping (AR+TS) at extreme values of aspect ratio (above 1.8 and below 0.7).

## 7. Evaluation

We compare our method to the other methods, by metric evaluations, by visual results, and through a user study. We compare against 4 methods:

- Scale and Object-Aware Saliency (SOAT): In this method scale and object-aware saliency is computed and a greedy algorithm is used to conduct region search over the generated saliency map [21].

- Efficient Cropping: This method generates a saliency map and conducts region search in linear time. Unlike previous methods, the search can be restricted to regions with a specific aspect ratio. However, when the aspect ratio mismatch between the image and the thumbnail is significant, a solution may not exist. In such cases, we apply the method without aspect ratio restriction. We run this method at a saliency threshold of 0.7. [2].

- Aesthetic Cropping: This method attempts to generate a crop with an optimum aesthetic quality [23].

- Visual Representativeness and Foreground Recognizability (VRFR): This method is very similar to ours, in objective. However, it can only generate thumbnails for a fixed size of $160 \times 120$ [9].

Since the code was not released for the aesthetic method and VRFR, our comparison is limited to a user study on their publicly available data set of 200 images with their generated thumbnails [9].

### 7.1. Metrics

For comparing different methods, we use the same metrics that were used in the experiments section. In addition, we use the hit ratio $h_r$ and the background ratio $b_r$ [9], which are defined as:

$$h_r = \frac{|g \cap p|}{|g|} \quad \text{and} \quad b_r = \frac{|p| - |g \cap p|}{|g|}$$

where $g$ is the ground truth box and $p$ is the predicted box. The metrics are computed over an annotated test set consisting of 7,005 annotations over 3,910 images and averaged. Table 4 shows the performance of different methods. Note that offset is higher than that reported in [9]. Unlike the MIRFLICKR-25000 dataset [10] that was used in [9], the data set we use has images with larger variation in size, some with low quality, and it includes many images with multiple objects. In addition, our thumbnails have an aspect ratio that varies from 0.5 to 2. This makes the data set significantly more challenging and would explain the large increase in the offset values compared to the reported results in [9]. We find that our method performs the best in terms of offset, rescaling factor, and IoU. We note that efficient cropping has a non-zero aspect ratio mismatch, indicating that there were examples where the problem was infeasible when the aspect ratio restriction is imposed. This is expected given the wide variation of thumbnail aspect ratios in our data set. Unsurprisingly, SOAT, which is agnostic to the target aspect ratio, has the highest aspect ratio mismatch.

The hit ratio represents the percentage of ground truth that was captured by the bounding box and the background ratio represents the percentage of bounding box area that lies outside the ground truth. The optimal method should be close to the ground truth and therefore should have a large hit ratio and a small background ratio. We find that the performance of different methods in terms of hit and background ratios is similar to the results reported in [9]. Namely, the saliency based methods focus on a relatively small region having large saliency. This leads to small crops which explain the low values for the hit and background ratios. Our method, in comparison is distinguished by a large hit ratio and a low background ratio. This shows that it closely matches the predicted ground truth boxes.

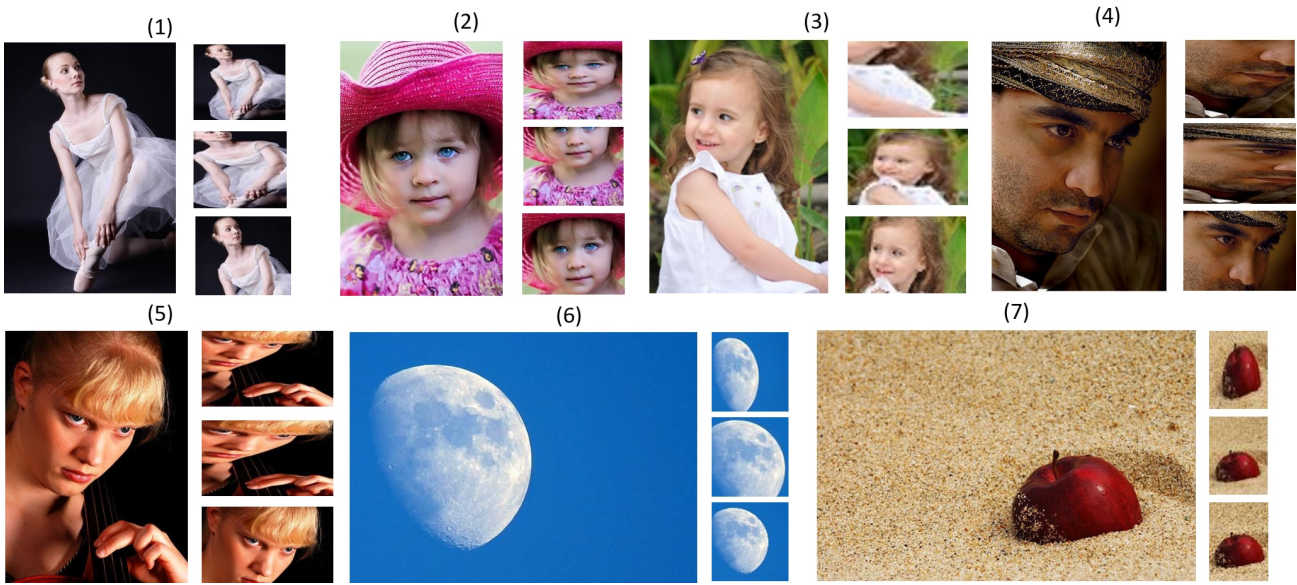

Figure 6. Images and their generated thumbnails: The original image is on the left, to its right we display the thumbnails: top is SOAT, middle is efficient cropping, and bottom is Fast-AT.

| Method | offset | rescaling factor | IoU | aspect ratio mismatch | $h_r$ | $b_r$ |
|---|---|---|---|---|---|---|
| SOAT [21] | 80.5 | 1.378 | 0.52 | 0.204 | 68.7% | 41.6% |
| Efficient Cropping [2] | 88.3 | 1.329 | 0.52 | 0.176 | 64.4% | 34.3% |
| Fast-AT (aspect ratio) | 55.0 | 1.148 | 0.68 | 0.010 | 83.7% | 37.1% |

Table 4. Metrics evaluated on different thumbnail generation methods.

## 7.2. Visual Results

We show qualitative results in comparison to other baselines in Figure 6. Saliency based methods succeed in preserving important content; in some examples however, their final thumbnails can have pronounced deformations. This can be seen in many examples for SOAT and in some examples for efficient cropping. In addition, these methods ignore the semantics of the scene and may ignore important parts of the image. This can be seen in the third and fourth examples for SOAT and in the first and second examples for efficient cropping. At the same time, it can be seen from the examples that Fast-AT succeeds in each case. It preservers the content of the scene and predicts thumbnails that tightly enclose the most representative part of the image.

## 7.3. User Study

We performed a user study where users were shown the original image and the generated thumbnails. They were asked to select the best thumbnail among SOAT, Efficient Cropping and Fast-AT. A total of 372 images were randomly picked from the test set. 30 mechanical turk users participated and no user was allowed to vote on more than 30 images. We have included the results of this study in Table 5. Fast-AT clearly outperforms the other two methods.

| SOAT [21] | Efficient Cropping [2] | Fast-AT |
|---|---|---|
| 88 (23.7%) | 86 (23.1%) | 198 (53.2%) |

Table 5. Number of votes for each method.

We performed another study over the released 200 images from [9] using the results of [21, 9, 23]. The results are shown in Table 6. Although VRFR[9] takes 60 seconds for an image and works for only one thumbnail size (160x120), Fast-AT performs slightly better in the user study performed.

| SOAT [21] | Aesthetic [23] | VRFR [9] | FastAT |
|---|---|---|---|
| 34(8.5%) | 92(23%) | 135(33.7%) | 139(34.7%) |

Table 6. Number of votes for each method.

## 8. Failure Cases and Multiple Predictions

We also investigate failure cases of Fast-AT. We look for examples in the test set where the prediction has an IoU with the ground truth below 0.1. Figure 7(a) shows some examples. The ground truth box is in green and the prediction is in blue. We see that although the prediction is very different from the ground truth, in some cases it still predicts crops that capture representative regions in the original images. Furthermore, for some of these failure cases we take the prediction with the second or third highest confidence. Figure 7(b) shows examples where the second or third predictions are close to the ground truth. This suggests that if the system is to be deployed, users could benefit if the system outputs a small set of top predictions instead of one. These predictions can be treated as a set of candidates from which the user picks the best solution. We also see a significant improvement in the performance of some metrics if the second prediction is also used. The third prediction did not lead to a significant improvement, as shown in Table 7.

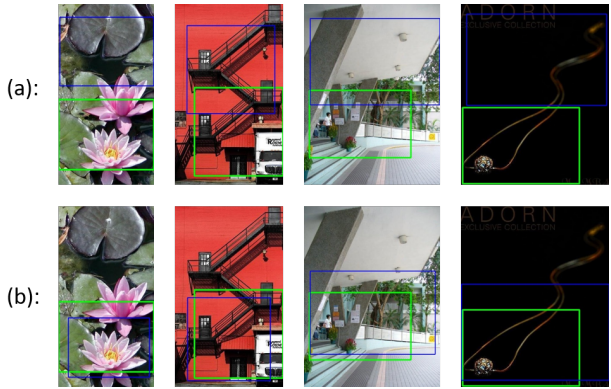

Figure 7. Failure cases of Fast-AT:(a): the prediction could have low IoU with the ground truth but still capture a representative region. (b): we show that the second or third most confident prediction is close to the ground truth.

| Model | offset | rescaling | IoU | mismatch |
|-------|--------|-----------|-----|----------|
| Top 1 | 55.0 | 1.149 | 0.677 | 0.010 |
| Top 2 | 50.4 | 1.152 | 0.693 | 0.011 |
| Top 3 | 50.3 | 1.152 | 0.693 | 0.011 |

Table 7. Performance of Fast-AT using the top 1, 2, and 3 predictions. The offset and IoU are significantly improved by using the top 2 predictions, the other metrics do not change significantly. Using the third prediction does not lead to significant improvement.

Another interesting case is when there is a significant aspect ratio mismatch between the representative part of the image and the thumbnail's aspect ratio. Because of the significant aspect ratio mismatch, the crop cannot capture all of the representative part of the image. We show that our algorithm is capable of producing multiple crops that cover different representative parts of the image. Figure 8 shows some examples. In the first three images (from the left), the region of interest is spread horizontally, but the thumbnail's aspect ratio is very small (tall thumbnail). The reverse is true for the rightmost image. The first row shows the bounding box prediction with the highest confidence and the second row shows the bounding box prediction with the second highest confidence. We see that these predictions cover different representative parts of the image.

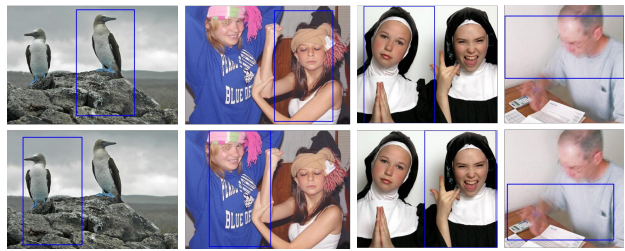

Figure 8. In the above images there is a significant aspect ratio mismatch between the region of interest in the image and the thumbnail's aspect ratio. The prediction with the highest confidence is shown in the first row and the prediction with the second highest confidence is shown in the second row. The predictions are successful in covering different representative regions in the image.

## 9. Conclusion

We presented a solution to the automatic thumbnail generation problem that does not depend on saliency or heuristic considerations but rather attacks the problem directly. A large data set consisting of 70,048 annotations over 28,064 images was collected. A CNN designed to generate thumbnails in real time was trained using this set. Metric and qualitative evaluations have shown superior performance over existing methods. In addition, a user study has shown that our method is preferred over other baselines.

## 10. Acknowledgement

We thank graduate students in our lab who annotated an initial dataset which was collected to check the feasibility of this approach. We are also grateful to the anonymous reviewers who provided valuable feedback for improving this paper.